\pdfoutput=1
\documentclass[11pt]{article}

\usepackage{microtype}
\usepackage{graphicx}
\usepackage{subfigure}
\usepackage{booktabs}
\usepackage{svg}
\usepackage{tikz}
\usepackage{amsmath}
\usepackage{amssymb}
\usepackage{mathtools}
\usepackage{amsthm}
\usepackage{acl}
\usepackage{times}
\usepackage{latexsym}
\usepackage[T1]{fontenc}
\usepackage[utf8]{inputenc}
\usepackage{microtype}
\title{Word-Level Representation From Bytes For Language Modeling}

\author{Chu-Tak Lee \\
  School of Computer Science \\
  Fudan University \\
  \texttt{lizd20@fudan.edu.cn} \\ \And
  Qipeng Guo \\
  Amazon AWS AI \\
  Amazon\\
  \texttt{gqipeng@amazon.com}\\ \And
  Xipeng Qiu \\
  School of Computer Science \\
  Fudan University \\
  \texttt{xpqiu@fudan.edu.cn} \\
}

\begin{document}
\maketitle 
\begin{abstract}

Modern language models mostly take sub-words as input, a design that balances the trade-off between vocabulary size, number of parameters, and performance. However, sub-word tokenization still has disadvantages like not being robust to noise and difficult to generalize to new languages. Also, the current trend of scaling up models reveals that larger models require larger embeddings but that makes parallelization hard.  Previous work on image classification proves splitting raw input into a sequence of chucks is a strong, model-agnostic inductive bias. Based on this observation, we rethink the existing character-aware method that takes character-level inputs but makes word-level sequence modeling and prediction. We overhaul this method by introducing a cross-attention network that builds word-level representation directly from bytes, and a sub-word level prediction based on word-level hidden states to avoid the time and space requirement of word-level prediction. With these two improvements combined, we have a token free model with slim input embeddings for downstream tasks. We name our method Byte2Word and perform evaluations on language modeling and text classification. Experiments show that Byte2Word is on par with the strong sub-word baseline BERT but only takes up 10\% of embedding size. We further test our method on synthetic noise and cross-lingual transfer and find it competitive to baseline methods on both settings.\footnote{Preprint.} 

\end{abstract}

\section{Introduction} 

Language models are the foundation of modern natural language processing. From Word2Vec \cite{word2vec} to BERT \cite{bert} and GPT-2 \cite{gpt2}, one key ingredient to make learning general language representation successful is switching from word-level \cite{word2vec, glove} input to subword-level \cite{allyouneed} input. Subword tokenization reduces the size of vocabulary, hence reducing the size of input and output embeddings, making training on a colossal size of text crawled from the Internet computationally tractable. However, reliance on sub-word level tokenization limits the capabilities of current NLP systems. Because the way of how subword tokenization algorithms work, subword language models are not robust to the variability noise of inputs. Typos, variably spelling, capitalization, and morphological changes can all cause the token representation of a word or phrase to change completely. What's more, recently Scaling Law \cite{scalinglaw} discloses larger network requires larger embedding, GPT-3 \cite{gpt3}, Bloom \cite{bloom} and XLM-R \cite{xlm-r} all of them have a whopping dictionary size of 250k. But larger embedding layer makes model parallelism hard since it tries to split the model into equal-sized chunks. 

ByT5 \cite{byt5} tries to alleviate these issues by switching to byte-level input. Due to the quadratic complexity of Transformer, byte-level models need more overhead to process the same input phrase(for example, MNLI usually requires a sequence length of 128 for subwords, for byte-level it is >1k), while still being inferior in performance. Charformer \cite{charformer} suggests using learnable, parametric tokenization. However this method does not recognize word boundaries so to shorten the sequence length, downsampling is naively shrinking the sequence after tokenization. And this method predicts the masked word by auto-regressive decoding, which add uncertainties when sampling and make them harder to train. 

\begin{figure*}
  \centering
  \includegraphics[scale=0.53]{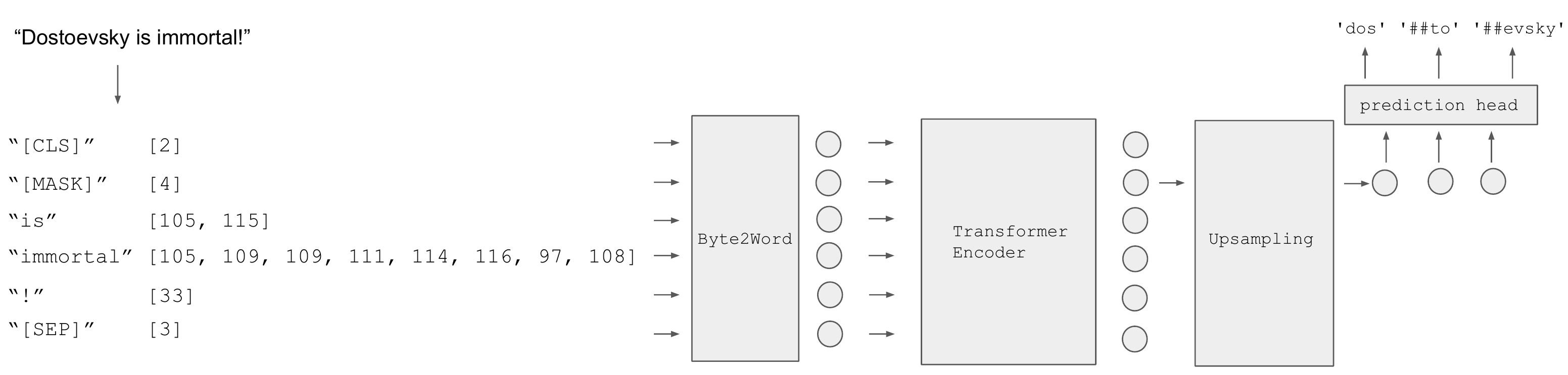} %
  \caption{Byte2Word}
  \label{fig:b2w}
\end{figure*}

Recent advances \cite{vit, convmixer} on image classification reveals that segmenting an image to patches boosts the performance of both transformer and convolution models. This phenomenon may indicate that segmentation and token boundary can act as crucial information for learning representation and urge us to revisit previously introduced character-aware methods \cite{charaware, limitsoflm, characterbert}. Character-aware methods process characters via a 1D convolution network with a suite of kernels with different widths to capture information of diverse granularities. After convolution, word-level hidden states are constructed by concatenation of length-wise max-pooled output signals. We modernize this process with a much more general cross-attention method. Character-aware methods predict words, so a reasonable-sized word-level vocabulary is needed for prediction to reduce compute budget. \cite{limitsoflm} proposes a contrastive learning method that generates word-level output embeddings on the fly. Also, we can decode the masked words in an auto-regressive style. Nevertheless, we do not adopt these methods in the mind of training stability and computational budget. We demonstrate that a simple yet effective way to train a word-level model without word-level vocabulary is to reuse a subword-level vocabulary by predicting subwords given masked, word-level hidden states from the last encoder layer. Seeing as subword-level vocabulary is only needed when pretraining and the subword-level output embeddings will be discarded for finetuning, what we obtain in the end is a token-free language model that processes sentences in bytes and with slim input embeddings, ready for transfer learning on downstream tasks. Our method is named Byte2Word and illustrated in Figure \ref{fig:b2w}. Through experiments on masked language modeling and downstream text classification, we show that Byte2Word is on-par with BERT on the GLUE benchmark \cite{glue}, proving that learning word-level representations from bytes can retain BERT-level performance while shrinking the embedding size by 10x. To show that our method is robust to synthetic noise, we inject four types of noise into the text of MNLI and find that our method is competitive to byte-level models while using much less computation. Besides, we find that our method has better cross-lingual transfer ability than subword methods.

\section{Related Work}

\noindent\textbf{Multilingual Language Modeling} XLM \cite{xlm} shows processing multiple languages with a shared vocabulary trained via Byte Pair Encoding \cite{bpe} improves the alignment of embedding spaces. However, this results in a 3x increase in vocabulary size compared to BERT. XLM-R \cite{xlm-r} aims to scale this setting with more parameters and languages, resulting in a lookup table that has a size of 250k and an embedding layer that takes up more than 47\% of total model parameters. RemBERT \cite{coupling} presents decoupled embeddings can allow the model to have more flexibility. In this sense, they rebalanced the input and output embedding of mBERT \cite{bert} and achieved better performance on multilingual language modeling than XLM-R, despite using a lot less of trained tokens.\\

\noindent\textbf{Word-level language modeling} While most transformer-based language models are built on top of subword tokenization, word-level transformer language models are not entirely infeasible. WordBERT \cite{wordbert} is the first word-level BERT that achieves better performance on cloze test, sequence labeling, and question answering compared to BERT. WordBERT utilizes negative sampling to successfully train a bidirectional transformer encoder with a vocabulary size of 1 million.\\

\noindent\textbf{Byte-level or character-level language modeling} For English-only tasks, byte-level and character-level tokenizations are equivalent because each English character takes only one byte if you ignore non-ASCII characters. But to incorporate other languages, character-level methods need to expand their vocabularies. Byte-level methods do not have this issue, but languages using some of the other scripts take more than one byte to represent a character. For example, Greek takes around 2 bytes and east Asian languages take around 3 bytes. This results in an even longer sequence length. Despite these restrictions, ByT5 \cite{byt5} shows a seq2seq transformer with re-balanced encoder and decoder depths plus more training that can achieve competitive performance on a diversity of tasks. CANNIE is a character level model that has a vocabulary of 1 million characters so it utilized hash embedding \cite{hashembedding} to curb large parameter amounts. Charformer \cite{charformer} proposes an attention-based, learnable tokenizer that builds subword embedding. But since Charformer has no concept of word boundaries, it can only reduce input length by naively downsampling. \\

\noindent\textbf{Character-aware method} Given a character sequence of a single word, previous work \cite{charaware, characterbert} use a 1D CNN network with multiple kernels to capture information of different granularities. Information extracted by the CNN is then pooled and concatenated to build the final word embeddings. \\

\noindent\textbf{Augmenting subword-level models with character-level infomation} CharBERT \cite{charbert} features a dual channels method that fuses subword information and character level information together and a character-level de-noise training that forces the model to reconstruct correct spelling of a word. Char2Subword \cite{char2suword} modifies a subword level Transfomer with a module that learns to reconstruct BERT subword embeddings from character-level inputs. Experiments show this method has superior performance on data with code-switching. Both of these methods are built on top of a pretrained model.

\section{Byte2Word}

Our goal of designing Byte2Word is to take an existing character-aware method and perform a set of improvements to make it token-free but still has on-par performance compared to subword-level models. These requirements emerge two challenges 1) How to construct word-level representations from bytes? 2) How to predict masked words without a word-level dictionary? We address these challenges in the following sections.

\subsection{Granularity of a Token}

Byte-level and word-level tokenization are two extremes when it comes to representing text into indices of a vocabulary. The simplest method, character-level tokenization is to let the vocabulary $V$ be the alphabet. This method is hard to practice on multilingual text as the total size of alphabets of multiple languages can be up to a million. One way to tackle this problem is switching to token-free byte-level input because there are only 256 bytes, which can represent text in the smallest level of granularity. However, this tends to yield very long sequences with sparse information. Word-level tokenization lies on the other end of the spectrum. Word-level tokenization tends to require a very large vocabulary to cover a diverse amount of text. The copra BERT trained on contains more than 10 millions unique words. This leads to a large embedding layer and intense softmax computation when predicting. Tailored methods such as negative sampling are used in \cite{wordbert} to train a word-level model successfully. Transformers mainly use subword tokenization right when it was first introduced. An easy way to explain subword tokenization is it uses a vocabulary containing a set of commonly occurring word segments like 'ing' and 'ly'. However, \cite{optimalvocab} and \cite{optimaltransport} show that hyperparameters of subword methods, especially the size of the dictionary, affect final performance. Thus, some models \cite{gpt3} allow a certain amount of redundancy for tokenization, resulting in varying granularities of token in its vocabulary and indirect learning.

This paper chooses to combine byte-level and word-level methods as proposed in \cite{charaware}. However, to truly achieve token-free, we do not want to maintain a word-level vocabulary, therefore we need to come up with some rules or methods to split text sentences into words to provide bytes and word boundaries to the model. For splitting words in a sentence, the boundary of a word is well-defined in English. We split phrases into words simply by white space, punctuation, camel case, and snake case after following the preprocessing procedure used in \cite{bert}. However, this is not the optimal choice for some non-English corpora, we leave this issue for future work. 

\subsection{Down-sampling}

After splitting a sentence to words, in order to construct word-level embedding from bytes for one of these words, we use cross attention mechanism proposed in \cite{allyouneed}, which is an effective and more general pooling mechanism compared to character-aware convolutions. Given a sentence and let the $i$th word of that sentence contains $L$ bytes, we use the corresponding byte-level embeddings $(e_{i1}^B, e_{i2}^B, ..., e_{iL}^B)$ from a byte-level embedding lookup table $E \in R^{256 \times d}$ to build the word embedding $e^W_i$ with a embedding size of $d$. We pack these byte-level embeddings together as key matrix $K_i$ and value matrix $V_i$ for cross-attention. Specifically, we have

\begin{equation}
K_i = \textrm{Concat}(e_{i1}^B, e_{i2}^B, ..., e_{iL}^B) \cdot W_k \\
\end{equation}
\begin{equation}
V_i = \textrm{Concat}(e_{i1}^B, e_{i2}^B, ..., e_{iL}^B) \cdot W_v \\
\end{equation}
\begin{equation}
e^W_i = \textrm{softmax} (\frac{Q_i K_i^T}{\sqrt{d}})V_i \\
\end{equation}

Here, $Q_i \in R^{1 \times d}$ denotes $i$th learnable word-query for position $i$ and $W_k \in R^{d \times d}, W_v \in R^{d \times d}$ are matrices for key and value projections. We find having multiple word-queries for different positions helps convergence. After pooling by cross-attention, word-level embeddings are processed by a position-wise feed-forward layer as the standard procedure. Our word-level embeddings $e^W_i$ is then added by residual connection and other types of embeddings. Followed by a layer normalization \cite{layernorm} and a linear projection, we obtain our final word-level hidden states $H^W_i$ for our encoder. This linear projection tactic resembles the Factorized embedding parametrization method introduced in \cite{albert} and the rebalanced embeddings in \cite{coupling}. 

\begin{equation}
H^W_i = \textrm{LN}((\textrm{FFN}(e^W_i) + e^W_i + e^{\textrm{pos}}_i + e^{\textrm{type}}_i)) \cdot W_{\textrm{proj}} \\
\end{equation}

where $e^{\textrm{pos}}$ is positional embedding, $e^{\textrm{type}}$ is token type embedding,  $W_{\textrm{proj}} \in R^{d \times d_{\textrm{encoder}}}$ projects the word-level embedding to match the hidden size of the backbone encoder.

\subsection{Up-sampling and Predictions}

\begin{table*}[!ht]
\centering
\small
\setlength{\tabcolsep}{.8mm}
\caption{Performance on GLUE test sets. Results of ByT5 and Charformer are excerpted from \cite{byt5, charformer}}
\vskip 0.15in
\begin{tabular}{@{}lcccccccccc@{}}
\toprule
                                           & \bf MNLI-m/mm & \bf QNLI & \bf QQP & \bf RTE & \bf SST-2 & \bf MRPC & \bf CoLA          & \bf STS-B      & \bf Avg. \\
\#metric                                   & Acc           & Acc      & F1      & Acc     & Acc       & F1       & Matthew's corr.   & Spearman corr. & &        \\
\#Examples                                 & 393k          & 105k     & 364k    & 2.5k    & 67k       & 3.7k     & 8.5k              & 7k             & &        \\
\midrule
$\textrm{BERT}_{\textrm{Large}}$(336M)     & 86.4/85.5     & 92.5     & 72.4    & 80.0    & 94.6      & 88.3     & 61.3              & 85.8           & 82.97     \\
\midrule
$\textrm{ByT5}_{\textrm{Small}} \textrm{(300M)}$ & \_     & \_     & \_    & \_    & \_      & \_     & \_   & \_  & 80.50 \\
$\textrm{Charformer}_{\textrm{Base}} \textrm{(203M)} $ & 82.6/82.7     & 89.0     & 88.8    & \_    & 91.6      & 91.1     & 42.6              & 85.3           & 81.40  \\
\midrule
Byte2Word (304M) & 86.1/85.6     & 92.6     & 72.0    & 76.3    & 95.2      & 89.0     & 59.1              & 87.8           & 82.63    \\
\bottomrule
\end{tabular}
\label{tab:glue_tasks}
\end{table*}

\begin{table}[!ht]
\centering
\small
\setlength{\tabcolsep}{.8mm}
\caption{Statistical information of how many bytes in a word in English Wikipedia \& BookCorpus}
\vskip 0.15in
\begin{tabular}{@{}lcccccccccc@{}}
\toprule
                   & \textbf{English Wikipedia + BookCorpus} \\
\midrule
Max byte length    & 8664 \\
Mode byte length   & 8 \\
Mean byte length   & 9.15 \\
Std of byte length & 5.60 \\
\bottomrule
\end{tabular}
\label{tab:data}
\end{table}

While with the method above, it is sufficient for language modeling. Still, predicting at word-level is computationally intense due to the large size of the lookup table. \cite{wordbert} showcases a receipt to train a word-level BERT by utilizing negative sampling. But such a method is not direct to implement in our setting since we don't even have a word-level vocabulary. \cite{limitsoflm} proposes a contrastive learning based method that dynamically generates output embedding given the character-level embedding sequence of the target word. But contrastive learning can easily lead to model collapse and requires extra caring to train. There is another way to train a fully vocabulary-independent model, \cite{limitsoflm, byt5, charformer} suggest predicting a word by auto-regressively decoding a sequence of characters given a word-level representation.  However, due to efficiency constraints, \cite{limitsoflm}'s character-level decoding is based on a pretrained then frozen word-level model.  We try to avoid these complex designs and instead choose to predict on sub-word level. \cite{coupling} shows that decoupling the hidden sizes of input and output embeddings can enhance performance. We believe it is feasible to take a further step and decouple input and output text granularity. By predicting subwords we can save a lot of effort and keep our model behave similarly to whole-word masking models and make it comparable to BERT baseline if we reuse its subword vocabulary. Besides, this method is token-free anyway when transfer learning on non-generative tasks. Subword-level features can be constructed by up-sampling from word-level hidden states. To keep things simple, we use a positional up-sampling method. Say we have the representation $H_i$ of the $i$th word, that word contains multiple subwords and we want to want to predict the $j$th one, we have

\begin{equation}
H_{ij} = H_i + P_j
\end{equation}

Where $P_j$ is a positional query for the $j$th subword of a word. We feed $H_{ij}$ into a prediction head for the final subword level prediction. We find that simply adding positional queries to the word-level representation respectively works better than more complex attention up-sampling \cite{settransformer}.

\section{Experiments}

\subsection{Model Setup and Pretraining}
\label{hyperparams}

While Byte2Word makes more sense on multilingual data, we evaluate our method in English due to budget constraint. We follow \cite{budgetbert} and pretrain our model on English Wikipedia and BookCorpus \cite{bookcorpus}. This corpora consists ~20GB of raw text, and ~10 million unique words, which should be enough to test if our method can learn a diversity of words. We provide some statistical information about the corpora in Table \ref{tab:data}. We follow the masking strategy in \cite{bert} but slightly modify the input format to use specific byte values for special tokens. Specifically we substitute "[PAD]" for HEX0 , "[UNK]" for HEX1, "[CLS]" for HEX2, "[SEP]" for HEX3 and "[MASK]" for HEX4. For instance, instead of "[CLS] a [MASK] sentence exmaple. [SEP]", Byte2Word takes the input as $\textrm{"\textbackslash x00 a \textbackslash x03 sentence example. \textbackslash x02"}$. In this way our model learns each special token from a single index, sparing extra hassle. Also to control memory usage, we limit the max amount of bytes a word can contain to 128. Similar to $\textrm{BERT}_{\textrm{Large}}$, we adopt a 24-layer transformer encoder with 16 attention heads and 1024 hidden sizes for our model backbone. To minimize computation overhead, we limit the size of Byte2Word embedding layer $d$ to 192. In this way, our model consume less than $0.2\%$ extra FLOPS per inference compared to $\textrm{BERT}_{\textrm{Large}}$. We also find that increasing the width of byte2word embedding slows convergence in ablation study in \ref{hiddensizes}. To ensure pretraining efficiency we follow \cite{roberta, budgetbert} and discard next-sentence prediction. We build our code base upon \cite{budgetbert} and train our Byte2Word model for 230k steps on 8x NVIDIA T4 16 GB for roughly a month.

\subsection{Downstream Evaluation and Analysis}

We test the performance of our model on GLUE benchmark \cite{glue}, the standard evaluation suite on language understanding for pretrained language model, and compare our result to subword-level baseline $\textrm{BERT}_{\textrm{Large}}$. During finetuning, instead of performing a grid search over sets of hyper-parameters, we use a fixed set of hyper-parameters across all tasks for each model. Table \ref{tab:glue_tasks} shows test set results of the GLUE benchmark. Our Byte2Word model performs on par with $\textrm{BERT}_{\textrm{Large}}$ on MNLI, QNLI and QQP, outperforms it on SST-2, MRPC and STS-B. However, our model has lower results on COLA and RTE. Overall, this amounts to a <0.5\% difference in the average score, proving that our method can reach on-par performance with sub-word level model, while being toke-free for transfer learning and can accept and adapt to new words.

\section{Analysis}

\subsection{Learning with Synthetic Noise}

To explore Byte2Word's ability to handle noisy input, similar to ByT5 \cite{byt5}, we test our method on learning synthetic noise injected during transfer learning.  We experiment four synthetic noising schemes listed below

\begin{itemize}
\item Random drop: 10\% of characters will be dropped in a sentence
\item Repetition: 20\% of characters will be repeated for 1-3 times(drawn uniformly).
\item Uppercase: All characters will be converted to uppercase
\item Random case: All characters will be converted to uppercase or lowercase randomly. This pattern is simulating Alternating caps\footnote{https://en.wikipedia.org/wiki/Alternating\_caps} existing on the Internet
\end{itemize}

These types of noise are added to training and evaluation data. We compare our method to vanilla byte-level model ByT5 and case-sensitive subword-level model BERT Cased on MNLI. For ByT5, we adopt a method introduced in EncT5 \cite{enct5} to finetune ByT5's encoder on MNLI but keep the training budget close to our baseline model. Table \ref{tab:noise} shows the test performance of byte-level, subword-level and our hybrid method Byte2word on MNLI. Byte2Word has superior performance on all types of noise compared to subword-level model. On random dropping, ByT5 Encoder has less performance loss, but note that ByT5 is trained on mC4 \cite{mt5}, a much larger multilingual dataset crawled from the Internet which contains noisy text. On Random case, it's no surprise to see our method perform terribly, due to the strategy of splitting camel case in the input. This result also shows that word boundary is critical for learning language representations. Given the result of injecting random case noise after word segmentation degenerate minimal performance in Table \ref{tab:noise}, we presume the performance would be much better if we make fewer assumptions on data preprocessing. It's also interesting to see that while Byte2Word model is pretrained using an uncased, subword-level vocabulary, this setting does not hinder finetuning on noisy text. Our Byte2word method can learn the noise pattern on the fly and has least degeneration on various noise types.

\begin{figure}
  \centering
  \includegraphics[scale=0.3]{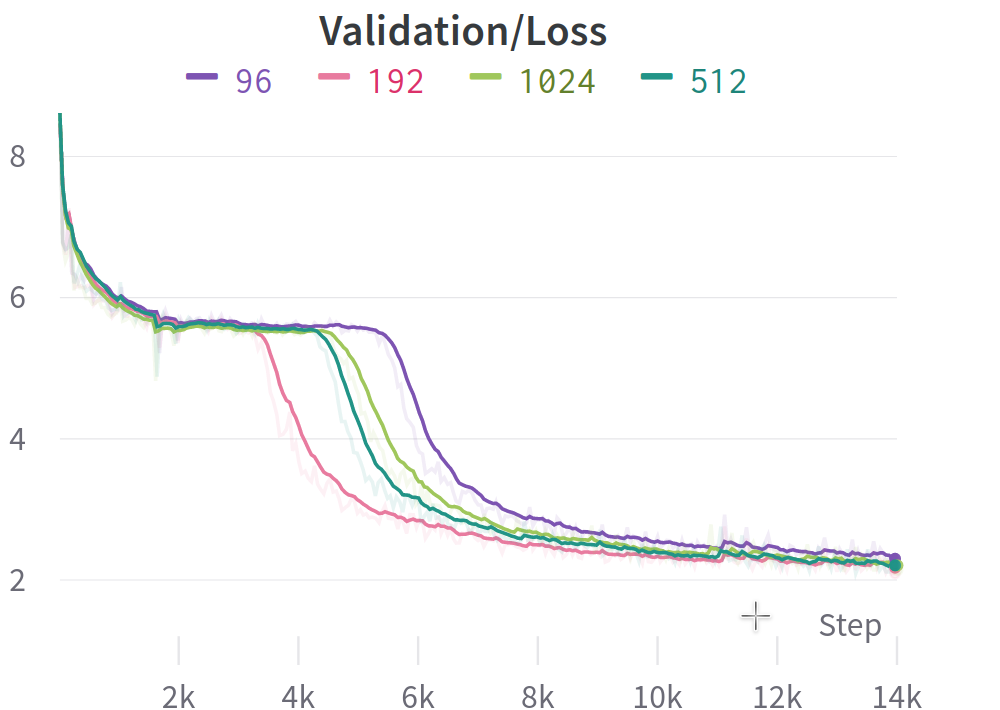} %
  \caption{Convergence rate of different embedding size}
  \label{fig:val_loss}
\end{figure}
\begin{table}[!ht]
\centering
\small
\setlength{\tabcolsep}{.8mm}
\caption{Test results on MNLI with synthetic noise}
\vskip 0.15in
\begin{tabular}{@{}lcccccccccc@{}}
\toprule
MNLI-m/mm   & Byte2Word           & BERT Cased & $\textrm{ByT5}_{\textrm{Base}}$ Encoder\\
\midrule
\bf Clean       & 86.1/85.6           & 86.4/85.7  &  84.2/83.0 \\
\midrule
\bf Random drop & 77.2/76.9           & 73.8/73.1  &  78.7/78.2 \\
\bf Repetition  & 85.8/85.2           & 78.8/78.7  &  81.7/81.4 \\
\bf Uppercase   & 86.5/85.8           & 77.9/77.2  &  83.2/83.0 \\
\bf Random case & 74.1/73.7           & 73.5/72.4  &  83.0/82.9 \\
\midrule
\bf Random case \\ \bf (post segmentation)   & 85.9/85.5        & \_  &  \_ \\
\bottomrule
\end{tabular}
\label{tab:noise}
\end{table}

\begin{table}[!ht]
\centering
\small
\setlength{\tabcolsep}{.8mm}
\caption{Test results on a subset of XNLI}
\vskip 0.15in
\begin{tabular}{@{}lcccccccccc@{}}
\toprule
Model                                        & \bf ar & \bf de  & \bf es & \bf fr & \bf ur \\
\midrule
$\textrm{mBERT}_{\textrm{Base}}$             & 72.59  & 76.95   & 78.66  & 77.96  & 62.55  \\
\midrule
$\textrm{BERT}_{\textrm{Large}}$ Uncased     & 64.99  & 71.62   & 74.11 & 75.01  & 56.4   \\
$\textrm{Byte2Word}_{\textrm{Large}}$        & 64.49  & 72.63   & 77.47 & 76.85  & 57.0   \\
\bottomrule
\end{tabular}
\label{tab:xnli}
\end{table}

\begin{table}[!ht]
\centering
\small
\setlength{\tabcolsep}{.8mm}
\caption{Spearman's correlation between cosine distance of learned words and edit distances}
\vskip 0.15in
\begin{tabular}{@{}lcccccccccc@{}}
\toprule
               & \bf levenshtein & \bf lcs & \bf jaro & \bf jaro-winkler \\
\midrule
Spearman corr. & 49.09           & 29.48   & 73.86    &  86.81           \\
\bottomrule
\end{tabular}
\label{tab:spearman}
\end{table}

\subsection{Cross-lingual Transfer}

Unlike \cite{byt5}, we evaluate cross-lingual transfer ability of our method by finetuning our English-only pretrained model on non-English downstream datasets. Since our model is not pretrained on multilingual corpora, it has to utilize cognates or shared grammar to facilitate learning. We test our model on XNLI \cite{xnli} which is basically MNLI translated into multiple languages. Our results on a subset of XNLI are shown in Table \ref{tab:xnli}. While the performance of our method is still behind multilingual language models, it has the better ability when adapting to new

\noindent languages compared to subword method. This result urge us to apply our method to multilingual language modeling in future work.

\subsection{Cosine Similarity Between Word Representations}

We analyze our learned Byte2Word model by calculating the cosine similarity of the learned word-level representations. Compare to BERT's sparse embedding space(Figure \ref{fig:cos}), we speculate representations learned by our Byte2Word method seems to contain less semantic meaning, as the cosine distance between "live" and "liver" is too close. Of course, hidden states usually are not as sparse as items of an embedding table. Based on these findings, we use word pairs from \cite{unimorph} and compute Spearman's correlation between cosine similarity of the learned representations and various types of edit distance of these pairs. Results are in Table \ref{tab:spearman}. We can see that our representations have a strong correlation with the jaro-winkler distance, which factors in matching characters and transpositions. We believe that a single layer of cross-attention with limited hidden size serves as an information bottleneck that makes representations only contain character-level, shallow abstraction and saves the high-level learning for the following encoder layers.  Also, it's interesting to point out that while no case information existed in supervised signal during pretraining, Byte2Word still can distinguish uppercase and lowercase in text.

\subsection{Convergence Rate of Different Embedding Sizes}
\label{hiddensizes}

To make sure that our choice of small embedding size $d$ for Byte2Word model does not degrade performance, we do an ablation study on the convergence rate of different embedding sizes. We use the previous setting in \ref{hyperparams} to pretrain Byte2Word models with multiple embedding sizes of 96, 192, 512, and 1024 but limit the total training step to 23k, roughly 10\% of our pertraining budget. The result is shown in Figure\ref{fig:val_loss}. We can see that while all models reach similar perplexity after a certain amount of computing, 192 has the fastest convergence rate. With the mind of limiting the hidden size of Byte2Word module as small as we can, choosing 192 seems like a great choice that balances computation and performance. 

\subsection{Comparison to Character-aware 1D CNN method}

To showcase our method is superior to previous, 1D convolution-based methods that consult the characters of a token to produce representation of a single word, we pretrain a Byte2Word model and CharacterBERT using the 10\% of our pretraining budget in \ref{hyperparams} and evaluate their performance on MNLI, the most representative task of the GLUE benchmark. Similar to \cite{char2suword}, we use BERT's vocabulary as a pseudo word-level vocabulary to minimize architecture differences compared to subword-level baseline. Results in Table \ref{tab:cnn} show that attention based pooling method of Byte2Word performs better than 1D CNN pooling.

\begin{table}[!ht]
\centering
\small
\setlength{\tabcolsep}{.8mm}
\caption{Dev results on MNLI}
\vskip 0.15in
\begin{tabular}{@{}lcccccccccc@{}}
\toprule
Model     & Embedding Layer Size & \bf MNLI Acc \\
\midrule
Subword   & 31M            & 83.77  \\
\midrule
CNN       & 0.4M           & 81.96  \\
Byte2Word & 0.8M           & 83.29  \\
\bottomrule
\end{tabular}
\label{tab:cnn}
\end{table}

\subsection{Parameters \& Efficiency}

$\textrm{BERT}_{\textrm{Large}}$'s subword lookup table has a size of 30522 and contributes roughly 31M of parameters, nearly 10\% of the total model size. On the contrary, our Byte2word method has a small size lookup table and a following cross-attention pooling, composing about 0.4M of parameters. While it has 2x more parameters than 1D CNN character-aware embedding method, compared to BERT it is negligible. However, embedding lookup albeit its size does not require computation, but our Byte2Word method has extra computation with a ceiling of 25 MFLOPS per inference if we set our byte-level hidden size to 192 and input sentence to 512 words.

\section{Limitations \& Future Work}

While matching BERT performance in many cases, Byte2Word slightly underperforms on CoLA and RTE of GLUE Benchmark. Those two tasks are both small-sized. We speculate text domain of these tasks is not covered by our pretrain corpora. In future work, it will be important to pretraining Byte2Word on corpora with more noise and languages, such as OpenWebText and mC4. Also it will be interesting to test fly Byte2Word on many-to-one translation with our Byte2Word embedding on encoder but language specific decoder. 

\section{Conclusion}

In this work, we present Byte2Word, an improved character-aware method that is token-free, lightweight and less compute heavy. On downstream task quality, Byte2Word is on-par with BERT that relies on WordPiece vocabulary. On handling noisy input, our method is much superior to subword-level models and competitive with vanilla byte-level models. At the same time, the computation efficiency of our method is on-par with subword-level models and way higher than vanilla byte-level methods. On transfering to another language, Byte2Word shows better performance than subword baseline BERT. These results suggest that our hybrid method might be the right blend of both worlds and may lay the path toward future NLP models that are efficient at processing varied text.



\bibliography{anthology}
\bibliographystyle{acl_natbib}
\clearpage

\appendix
\onecolumn

\section{Appendix}
\label{sec:appendix}

\subsection{Hyperparameters}

Below we provide a complete list of hyperparameters for pretraining and finetuning on GLUE benchmark.

\begin{table}[!ht]
\centering
\small
\setlength{\tabcolsep}{.8mm}
\vskip 0.15in
\caption{Hyperparamters for pretraining Byte2Word BERT on English Wikipedia + BookCorpus}
\begin{tabular}{@{}lcccccccccc@{}}
\toprule
\bf Parameter & \bf Value \\
\midrule
Number of Layers    & 24   \\
Hidden size         & 1024 \\
Attention heads     & 16   \\
Attention head size & 64   \\
Byte2Word embedding layer size & 192 \\
Byte-level embedding size & 192 \\
FFN intermediate size & 4096 \\
Dropout             & 0.1  \\
Attention Dropout   & 0.1  \\
Layer norm type     & pre LN  \\
Learning Rate Decay & Linear\\
Weight Decay        & 0.01  \\
Optimizer           & AdamW \\
Adam $\theta$       & 1e-6  \\
Adam $\beta_1$          & 0.9   \\
Adam $\beta_2$          & 0.98  \\
Gradient Clipping   & 0.0   \\
Batch Size          & 4032  \\
Learning Rate       & 1e-3  \\
Warmup Proportion   & 0.06  \\
Max Steps           & 240K  \\
Max Length          & 128   \\
Prediction dictionary & BERT uncased \\
\bottomrule
\end{tabular}
\label{tab:pretrain_hyperparams}
\end{table}

\begin{table}[!ht]
\centering
\small
\setlength{\tabcolsep}{.8mm}
\vskip 0.15in
\caption{Hyperparamters for finetuning on GLUE benchmark}
\begin{tabular}{@{}lcccccccccc@{}}
\toprule
\bf Parameter                        & \bf $\textrm{BERT}_{\textrm{Byte2Word}}$ & \bf BERT & \bf ByT5 Encoder \\
\midrule
Batch Size                           & 32            & 32       & 16   \\
Weight Decay                         & 0.01          & 0        & 0    \\
Max Gradient Norm                    & 1.0           & 1.0      & 1.0   \\
Learning Rate                        & 5e-5          & 2e-5     & 8e-5 \\
{Max Epoch (MNLI, QQP, QNLI, SST-2)} & 3             & 3        & 3 \\
{Max Epoch (RTE, CoLA, STS-B)}       & 5             & 5        & \_ \\
Warmup Ratio                         & 0.001         & 0.       & 0.1 \\
Learning Rate Decay                  & Linear        & Linear   & Linear \\
\bottomrule
\end{tabular}
\label{tab:glue_hyperparams}
\end{table}

\subsection{Cosine Similarity}

\begin{figure}
  \centering
  \includegraphics[scale=0.25]{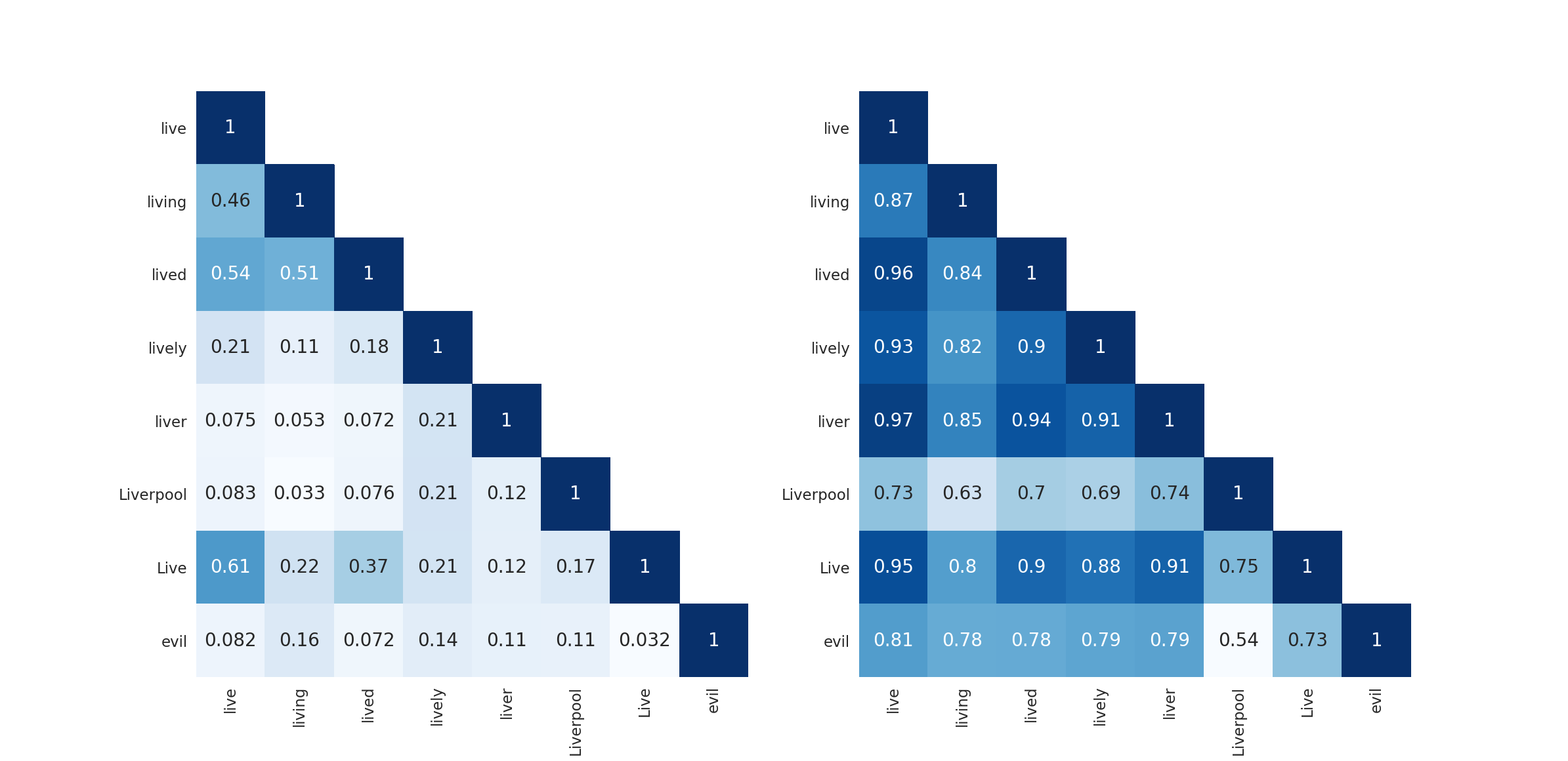} %
  \caption{Cosine similarity of $\textrm{BERT}_{\textrm{Large}}$ uncased(left) and our Byte2Word(right) embeddings}
  \label{fig:cos}
\end{figure}

\end{document}